\documentclass[runningheads]{llncs}
\usepackage[T1]{fontenc}
\usepackage{graphicx}
\usepackage{orcidlink}
\usepackage{amssymb}
\usepackage{booktabs}
\usepackage{multirow}
\usepackage{array}
\usepackage{marvosym}
\begin{document}
%
%\title{Geometric Morphometrics Benchmark for the Evaluation of Low-Cost 3D Facial Models}
\title{A 3D Facial Reconstruction Evaluation Methodology: Comparing Smartphone Scans with Deep Learning Based Methods Using Geometry and Morphometry Criteria}
%\title{A Geometric and Morphometric Methodology for Evaluating Low-Cost 3D Facial Acquisition and Reconstruction Techniques}
%
%
%\titlerunning{Abbreviated paper title}
% If the paper title is too long for the running head, you can set
% an abbreviated paper title here
%
\author{Álvaro Heredia-Lidón\inst{1}\textsuperscript{(\Letter)}\orcidlink{0000-0001-6325-8744} \and Alejandro Moñux-Bernal\inst{1} \and Alejandro González\inst{1}\orcidlink{0000-0002-3644-491X} \and Luis M. Echeverry-Quiceno\inst{2}\orcidlink{0000-0002-9799-0252} \and Max Rubert\inst{2} \and Aroa Casado\inst{2}\orcidlink{0000-0001-9116-9167} \and María Esther Esteban\inst{2}\orcidlink{0000-0002-2410-9700} \and Mireia Andreu-Montoriol\inst{2} \and Susanna Gallardo\inst{2} \and Cristina Ruffo\inst{2} \and Neus Martínez-Abadías\inst{2}\orcidlink{0000-0003-3061-2123} \and Xavier Sevillano\inst{1}\orcidlink{0000-0002-6209-3033}}
\authorrunning{}
\titlerunning{}

% First names are abbreviated in the running head.
% If there are more than two authors, 'et al.' is used.
%
\institute{HER, La Salle - Universitat Ramon Llull, Barcelona, Spain \\
\email{alvaro.heredia@salle.url.edu} \\
 \and
BEECA, Facultat de Biologia, Universitat de Barcelona, Barcelona, Spain\\}
\maketitle              % typeset the header of the contribution
\begin{abstract}
Three-dimensional (3D) facial shape analysis has gained interest due to its potential clinical applications. However, the high cost of advanced 3D facial acquisition systems limits their widespread use, driving the development of low-cost acquisition and reconstruction methods. This study introduces a novel evaluation methodology that goes beyond traditional geometry-based benchmarks by integrating morphometric shape analysis techniques, providing a statistical framework for assessing facial morphology preservation. As a case study, we compare smartphone-based 3D scans with state-of-the-art deep learning reconstruction methods from 2D images, using high-end stereophotogrammetry models as ground truth. This methodology enables a quantitative assessment of global and local shape differences, offering a biologically meaningful validation approach for low-cost 3D facial acquisition and reconstruction techniques. 

%Facial dysmorphologies appear as crucial diagnostic biomarkers in rare, psychotic and neurodevelopmental disorders. The analysis of facial dysmorphologies with diagnostic purposes still largely relies on qualitative visual assessments and basic anthropometric measurements, highlighting the need for quantitative low-cost 3D facial assessment tools that can be applied in clinical practice. Using stereophotogrammetry 3D facial reconstructions as the gold standard and Geometric Morphometrics as the facial biomarker computation framework, we present a comparative study between low-cost 3D facial reconstruction methods based on a 2D images and on scans obtained via mobile phone infrared cameras. Experimental results reveal that mobile scan reconstructions yield the most similar results to those of stereophotogrammetry in terms of biomarkers that quantify facial shape at local scale, while multi-image Hierarchical Representation Networks yield the most accurate biomarkers at global facial shape scale.

\end{abstract}

\section{Introduction}
\label{sec:intro}

Three-dimensional (3D) facial model acquisition has been a breakthrough in fields such as plastic surgery \cite{Li2016}, orthodontic diagnosis \cite{Sarver2015} or the study of facial morphology for diagnostic purposes \cite{Starbuck2021}. The potential of these techniques lies in their non-invasive nature, capturing precise details of the face, and resolving the limitations of two-dimensional (2D) analysis.

Acquiring highly accurate 3D facial models usually requires specialised equipment \cite{Parsa2022}. Examples include stereophotogrammetry systems (SPG) --consisting of multiple cameras capturing images synchronously \cite{Heike2010}--, portable devices based on laser technology \cite{Koban2020}, structural light-based cameras \cite{Li2017} or medical devices like magnetic resonance image scanners \cite{Heredia-Lidon2023}. While these techniques yield high-resolution 3D facial reconstructions, they are often expensive and inaccessible, limiting their widespread application in clinical settings \cite{Gibelli2018}.

As an alternative, portable structured-light scanners embedded in consumer devices like smartphones have emerged as a promising approach to low-cost 3D facial model acquisition \cite{DEttorre2022,Face3DBiomark}. Several studies have validated the applicability of the iPhone’s TrueDepth camera as a viable alternative to high-cost scanners for clinical applications \cite{DEttorre2022,Ritschl2024}.

A different low-cost approach consists in deep learning-based methods for reconstructing 3D facial models from a single or multiple 2D images. On the one hand, single-view 3D facial reconstruction has been a long-standing challenge in computer vision, leading to the development of various state-of-the-art approaches. For instance, Lei \textit{et al.} \cite{Lei2023} introduced a Hierarchical Representation Network (HRN) that decouples facial geometry to enhance high-fidelity detail reconstruction. Deng \textit{et al.} \cite{Deng2019} proposed a weakly-supervised CNN for 3D face reconstruction, while Wang \textit{et al.} \cite{Wang2024} leveraged geometric information from facial part silhouettes to guide the reconstruction process, improving structural accuracy.

\begin{figure}[!t]
\begin{center}
\includegraphics[scale=0.8]{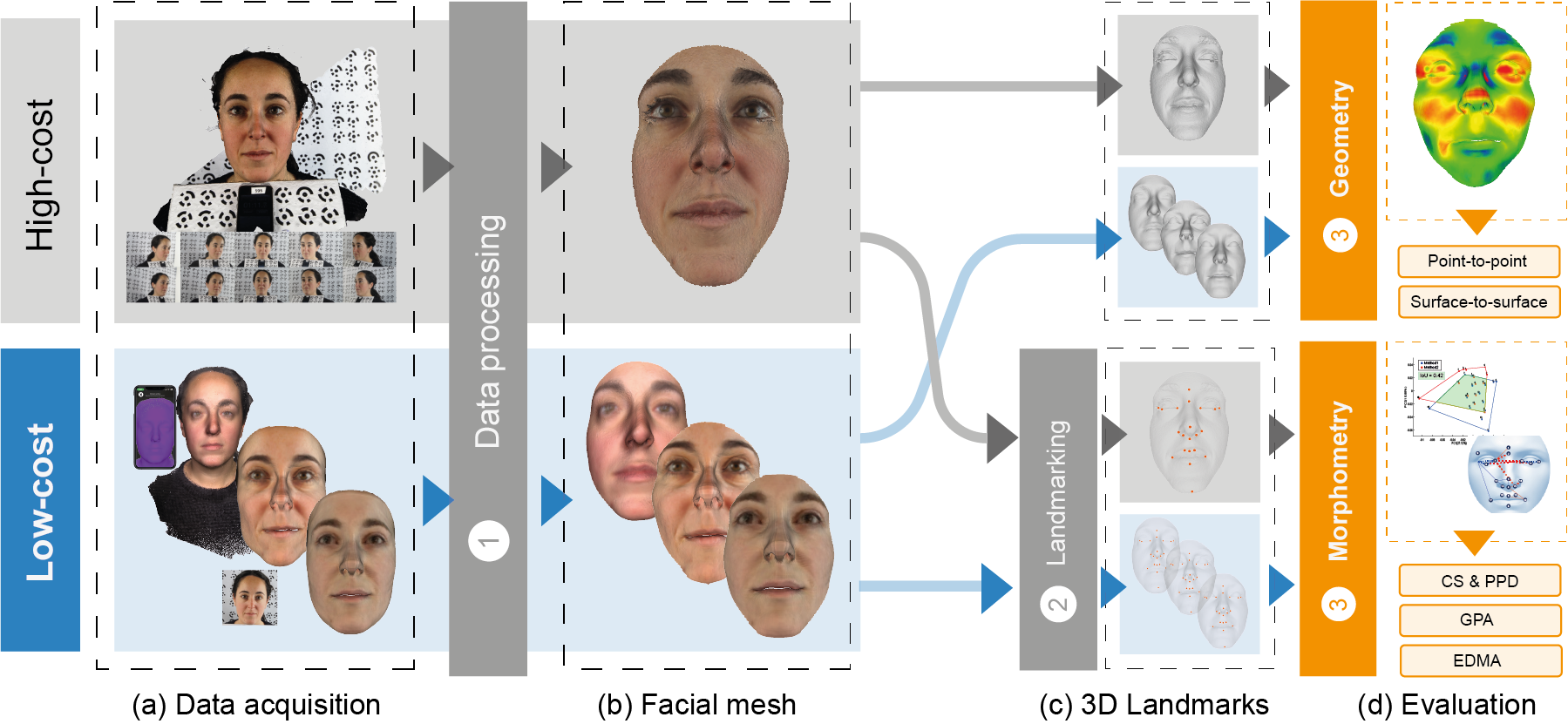}
\end{center}
\vspace{-0.5cm}
\caption{Methodology pipeline. (a) Data acquisition using high-cost 3D stereophotogrammetry system, and low-cost approaches. (b) Facial mesh extraction and normalisation. (c) 3D anatomical landmarks registration for morphometric analysis. (d) Methods comparison based on geometry and morphometry metrics.}
\label{fig:intro}
\end{figure}

On the other hand, multi-view methods emerge as a powerful alternative, offering more accurate modelling but relying on camera calibrations and poses \cite{Lei2023,Ramon2021}. Ramon \textit{et al.} \cite{Ramon2021} presented H3D-Net architecture for high-quality 3D human heads reconstruction from a few input images. Lei \textit{et al.} \cite{Lei2023} took advantage of the hierarchical structure of HRN to adapt it to multiple images, adding geometric consistency between different views. Another multi-view approach is diffusion modelling, generating multiple views from a single 2D image, like in Era3D \cite{Li2024}, for subsequent reconstruction of 3D models using NeuS \cite{Liu2024} or Gaussian Splatting \cite{Tang2024} techniques.

%Despite these advances, the validation of these low-cost reconstruction methods typically relies in geometric accuracy of the 3D facial reconstructions is typically measured in terms of metrics like Chamfer Distance, point-to-surface distance, inter-landmarks error, Mean Normal Error or region-averaged Normalized Mean Square Error with respect to some high-resolution 3D facial model used as the ground truth \cite{Lei2023} on widely used benchmarks such as REALY \cite{Chai2022} or NoW \cite{Sanyal2019}, where the geometric accuracy of the 3D facial reconstructions is typically measured in terms of metrics like Chamfer Distance, point-to-surface distance, inter-landmarks error, Mean Normal Error or region-averaged Normalized Mean Square Error with respect to some high-resolution 3D facial model used as the ground truth \cite{Lei2023}. However, from a clinical applicability standpoint, it would be much more relevant to investigate whether these 3D facial reconstructions preserve and reproduce the real facial anatomy using statistical analysis techniques.

The validation of low-cost 3D facial acquisition and reconstruction methods has predominantly relied on benchmarks such as REALY \cite{Chai2022} or NoW \cite{Sanyal2019}, which assess geometric accuracy using high-resolution 3D facial models as ground truth \cite{Lei2023}. However, these benchmarks are limited to surface-based geometric deviations and fail to account for facial morphology and anatomical consistency. To address this gap, it is necessary to incorporate statistical shape analysis techniques that evaluate the morphological fidelity of reconstructed faces, providing a more anatomically relevant validation framework.

In this context, Geometric Morphometrics (GM) is a robust set of statistical tools that measure the shape of biological structures from the coordinates of a set of anatomical landmarks \cite{Mitteroecker2009}. The main advantage of GM is that the 3D shape and geometry are preserved throughout the analysis, providing more robust and precise results, detecting subtle shape differences, and allowing the visualization of the results as shape changes \cite{Mitteroecker2009}. Among the two most commonly used GM techniques in facial analysis are: \textit{i)} Generalized Procrustes Analysis (GPA) \cite{Mitteroecker2009} to assess global differences in facial shape, and \textit{ii)} Euclidean Distance Matrix Analysis (EDMA) \cite{Lele2001} to detect local differences in facial shape. %These techniques can be boosted by the use of Statistical Shape Models (SSM), obtaining a detailed static analysis from Principal Component Analysis (PCA) \cite{Woods2017}.

For this reason, we propose a new evaluation methodology for low-cost 3D facial acquisition and reconstruction methods, which extends beyond traditional geometric assessments by incorporating GM-based metrics (Figure \ref{fig:intro}). To test our methodology, we compared smartphone-based 3D scans with 2D image-based deep learning 3D reconstruction methods using high-resolution stereophotogrammetry models as ground truth. The results showed that smartphone scans achieved higher geometric and morphometric similarity with respect to the ground truth. 

\section{Materials and Methods}

\subsection{Participants Recruitment and Ethical Statement}

A total of 82 volunteers (40 females and 42 males, ages 30.48 $\pm$ 12.13) with no pathological facial dysmorphologies were recruited. Each individual was captured on the same day by the different acquisition methods. No discriminatory criteria, such as excessive wrinkles or beard, were selected. To guarantee unobstructed facial shape acquisition, participants were asked to remove any hair covering their face and any artefacts such as earrings or piercings when possible, while maintaining a neutral facial expression. All the participants signed an informed consent approved by the Bioethics Committe of Universitat de Barcelona (IRB03099). 

\subsection{Face Acquisition and Reconstruction}

\begin{figure}[t]
\begin{center}
\includegraphics[scale=0.7]{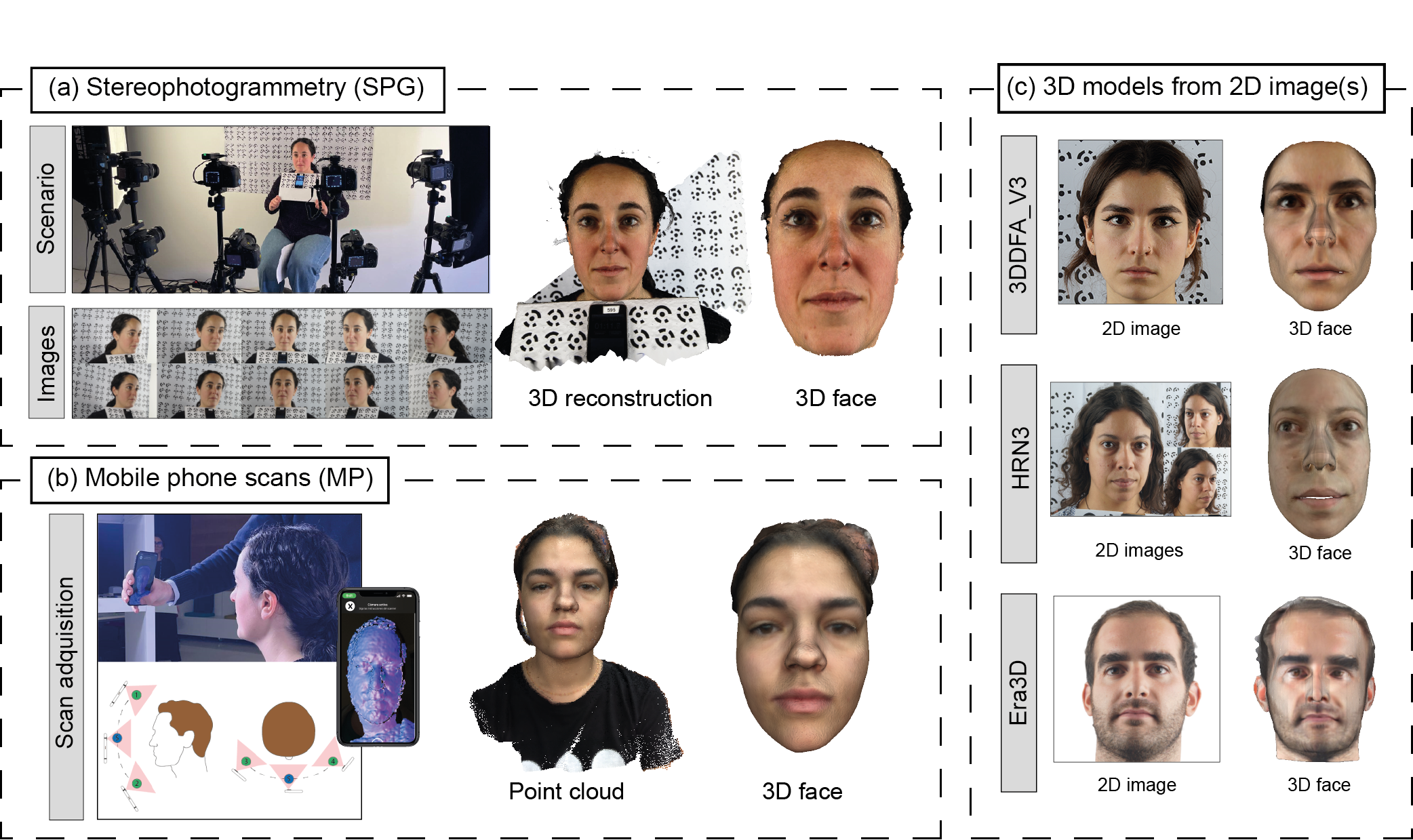}
\end{center}
\vspace{-0.5cm}
\caption{Face acquisition and reconstruction. (a) High quality facial models acquired by a stereophotogrammetry (SPG) system: setup and reconstruction of the 3D facial model from 10 images. (b) Low-cost facial model acquisition via iPhoneX TrueDepth camera and processed 3D model. (c) Low-cost facial reconstruction from 2D image(s): 3DDFA\_V3 \cite{Wang2024}, HRN3 \cite{Lei2023} (3 images) and Era3D \cite{Li2024}.}
\label{fig:methodology}
\end{figure}

The first step of the proposed methodology requires using a high cost acquisition method to obtain the ground truth facial models, and a set of methods to be compared against this ground truth. Figure \ref{fig:methodology} illustrates the implementation of this step in the context of this work.%The 3D facial model acquisition and reconstruction methods employed are outlined below ().

\paragraph{\textbf{High-end facial acquisition}.} We built a stereophotogrammetry (SPG) system composed of 10 synchronised DSLR cameras mounted on 5 tripods that were arranged in a semi-circle in two rows at different angles and heights covering on the front and sides of the subject (from 0 to $\pm$90 degrees), and two side light sources (Figure \ref{fig:methodology}a) as in \cite{Starbuck2021}. Camera synchronization was crucial to minimize motion artefacts \cite{Heike2010}. High-resolution 3D models were subsequently reconstructed and marker-based scaled from the 10 images set of each subject using the Agisoft Metashape software. The resulting 3D facial meshes were manually processed using MeshLab for cleaning, orientation normalization, and facial mesh extraction. 

\paragraph{\textbf{Low-cost acquisition/reconstruction.}} In this work, we applied the methodology to evaluate the following low-cost 3D facial acquisition methods:

\begin{itemize}
    \item {\textit{\textbf{Mobile phone (MP) scans.}}} We used Face3DBiomark \cite{Face3DBiomark}, a client-server architecture that uses an iPhoneX for 3D facial scanning through the TrueDepth infrared camera and a cloud server for mesh processing (Figure \ref{fig:methodology}b). The acquired facial 3D point-cloud was automatically processed with PyMeshLab for facial segmentation, orientation normalisation and remeshing. The acquisition of a facial scan took approximately 15 seconds at a distance of 25-50 cm from the subject.

    \item{\textit{\textbf{Facial reconstructions from 2D images.}}} We selected three of the most recent and accurate open-source methods currently available in the literature to reconstruct 3D facial models from 2D images obtained by the SPG system (Figure \ref{fig:methodology}c): 3DDFA\_V3 \cite{Wang2024}, HRN \cite{Lei2023} and Era3D \cite{Li2024}. For 3DDFA\_V3 and Era3D we used a frontal cropped image of dimension 256x256 pixels. In Era3D, background removal was performed to enhance accuracy. For HRN, a multi-view configuration using a frontal and two lateral views was employed, so we refer to the method as HRN3 hereafter. The resulting 3D facial meshes were processed using PyMeshLab for orientation and scale adjustments.
\end{itemize}

To ensure a fine morphological comparison, all low-cost reconstructions were manually aligned with their high-cost SPG counterparts using MeshLab’s Align Tool, based on five manually selected landmarks (inner eye commissures, nasal septum base, and labial commissure corners). The facial region of all reconstructions was extracted using a 100 mm sphere centered at the nose tip, and Polygon File Format (.PLY) was used for data storage.

\subsection{Automatic landmarking}

The next step of our methodology consists in establishing a standardized landmark-based representation of facial morphology for morphometric evaluations based on GM. 

To this end, an automatic and accurate facial landmarking method is required. In this work, 3D facial landmarking was performed by training a 21-landmark multi-view consensus CNN (MV-CNN) model \cite{Paulsen2019} on a proprietary dataset of 125 manually landmarked facial 3D models obtained from SPG scans, distinct from the evaluation dataset. For training hyperparameters, we set a 25-view configuration and the depth channel (\textit{z-buffer}) as the input image modality, following \cite{HerediaLidon2024}.

%To quantitatively assess facial shape using Geometric Morphometrics (GM) techniques, it is essential to establish a standardized landmark-based representation of facial morphology. This was achieved by recording the 3D Cartesian coordinates of 21 anatomical landmarks, as in \cite{Starbuck2021}.

%Since manual landmarking is labor-intensive, error-prone, and subject to inter-observer variability \cite{HerediaLidon2024}, we used the multi-view consensus CNN (MV-CNN) model \cite{Paulsen2019} for automatic 3D facial landmarking. A proprietary dataset of 125 manually landmarked facial 3D models obtained from SPG scans was employed, distinct from the evaluation dataset. For training hyperparameters, we set a 25-view configuration and the depth channel (\textit{z-buffer}) as the input image modality. Inferred landmarks coordinates were obtained as the average of 10 predictions with a maximum acceptable RANSAC value of 10, following \cite{HerediaLidon2024}.

\subsection{Evaluation metrics}

\begin{figure}[!t]
    \begin{center}
    \includegraphics[scale=0.8]{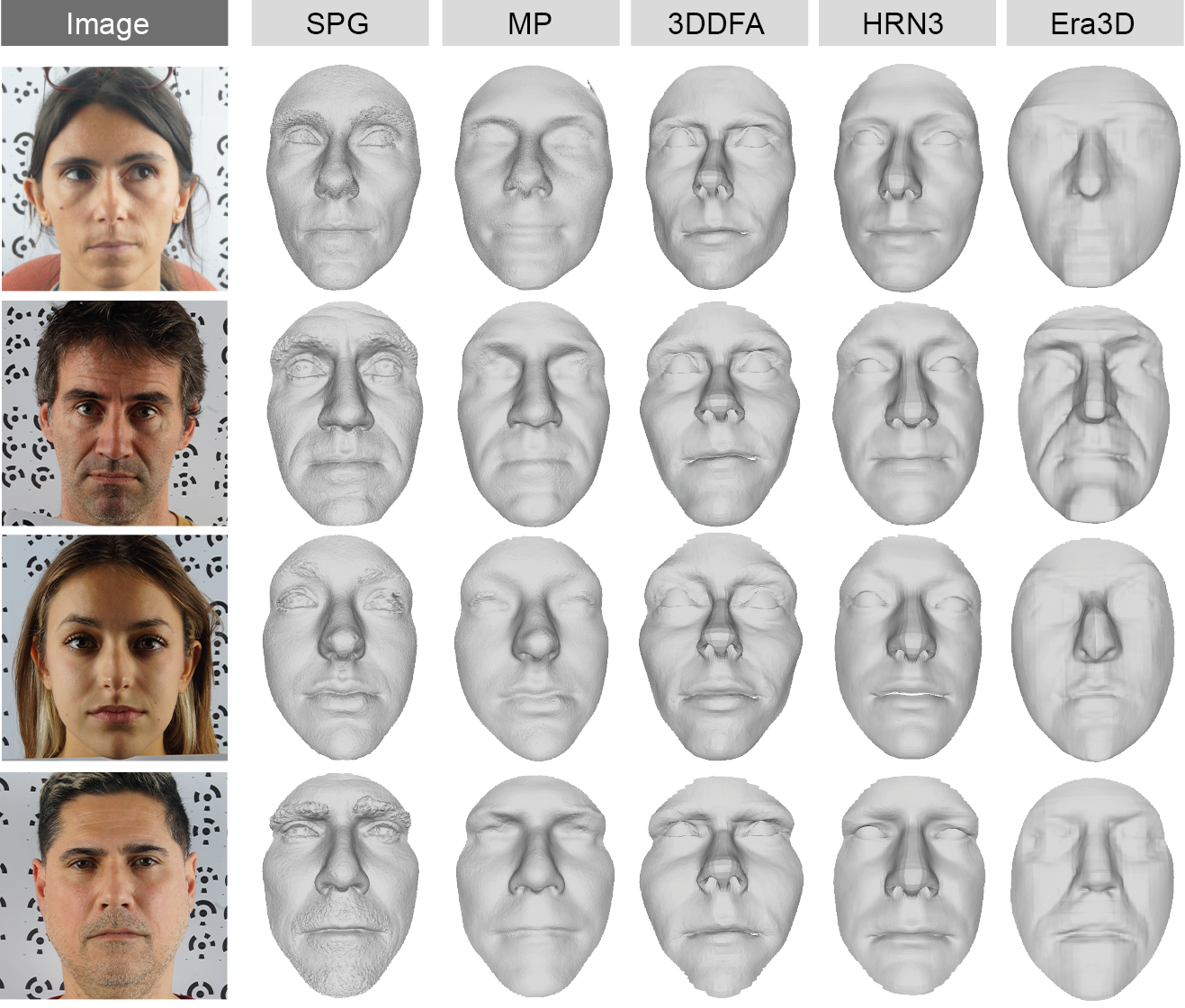}
    \end{center}
    \vspace{-0.5cm}
    \caption{At-a-glance comparison of the geometry of the 3D facial acquisition and reconstruction methods.}
    \label{fig:qualitative}
\end{figure}

Our methodology includes evaluation following two complementary approaches: \textit{i)} Geometric analysis –-measuring surface accuracy-- and \textit{ii)} GM-based morphometric analysis –-assessing facial morphology preservation.

\paragraph{\textbf{Geometric evaluation metrics.}} Point-to-point distance and surface-to-surface deviations are computed by comparing the low-cost reconstructions against their gold-standard equivalents. Since mesh vertex densities differed, point-to-point distances were computed using the nearest neighbour approach via kD-tree search.

\paragraph{\textbf{Morphometric evaluation metrics.}} This evaluation comprises quantifying facial morphology using Centroid Size (CS) and Pairwise Procrustes Distances (PPD) morphometrics variables \cite{HerediaLidon2024}, alongside Generalized Procrustes Analysis (GPA) and Euclidean Distance Matrix Analysis (EDMA) for global and local shape characterization.

%First, we assessed the correlation between SPG (gold-standard) landmarks and those from low-cost methods. CS was computed as the square root of the sum of squared distances from each landmark to the centroid, while PPD was derived from GPA, computing the Procrustes distances between each subject landmarks and the rest of the sample

To apply GPA, we performed Principal Component Analysis (PCA) to visualize global shape variation in morphospace, computing the Intersection-over-Union (IoU) between the convex hulls of high- and low-cost methods and calculated Procrustes Distance (PD) to quantify overall shape differences, with 10,000 permutations for statistical validation.

For EDMA, we analyzed 210 inter-landmark distances, comparing male and female subgroups within each reconstruction method. Using a confidence interval ($\alpha = 0.10$) \cite{Lele2001}, we identified $n$ significant inter-landmarks distances. To determine which method best preserved facial morphology, we computed the percentage of \textit{matching distances} (MD, \%) by comparing the top-5 ($n=5$) and top-10 ($n=10$) longest/shortest inter-landmark distances across methods.

\begin{figure}[t!]
    \begin{center}
    \includegraphics[scale=0.65]{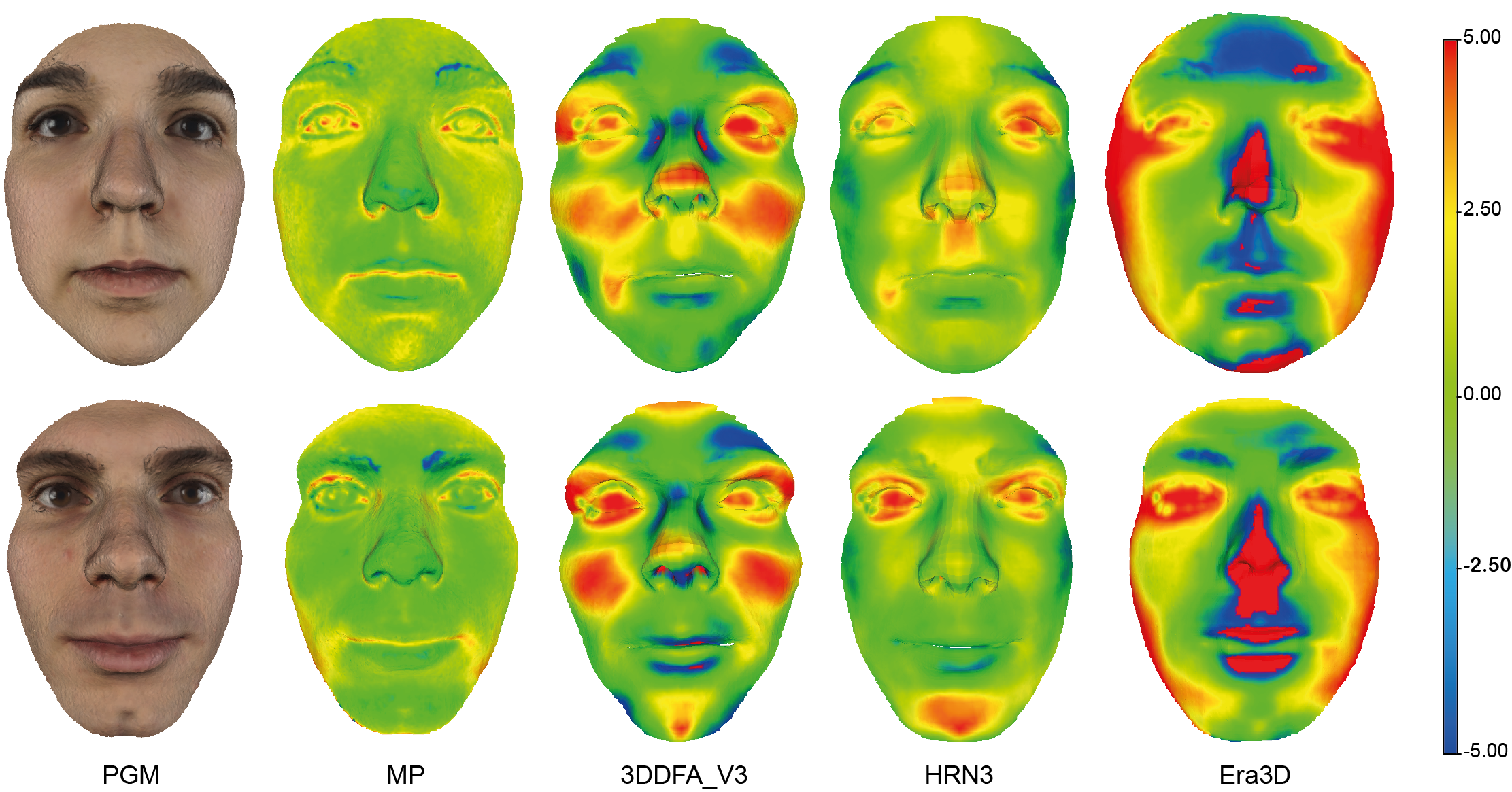}
    \end{center}
    \vspace{-0.5cm}
    \caption{Surface-to-surface desviation between low-cost methods and SPG. Deviation values are coded with a colour map from blue to red.}
    \label{fig:surface2surface}
\end{figure}

\section{Results and discussion}

\paragraph{\textbf{Experimental setup.}} All analysis results and 3D processing of the models have been performed on a Windows PC (16GB RAM with 6-core CPU). 3DDFA\_V3 and HRN3 models were generated on a Linux server (36-core CPU, NVIDIA GeForce RTX 2080 Ti GPUs) and Eras3D models with NVIDIA H100 GPUs provided by EuroHPC JU and MareNostrum5 at Barcelona Supercomputing Center. The code implementing all the steps of the methodology, the facial landmarks obtained on each facial model and the MVCNN automatic landmarking model are available at: \url{https://bitbucket.org/cv_her_lasalle/gmfacialbenchmark}.

\subsection{Geometric evaluation}

Figure \ref{fig:qualitative} illustrates examples of the 3D facial models obtained from the acquisition and reconstruction methods in our testbed. To focus on geometric accuracy, texture and vertex colours were omitted.

\begin{figure} [t]
    \begin{center}
    \includegraphics[scale=0.7]{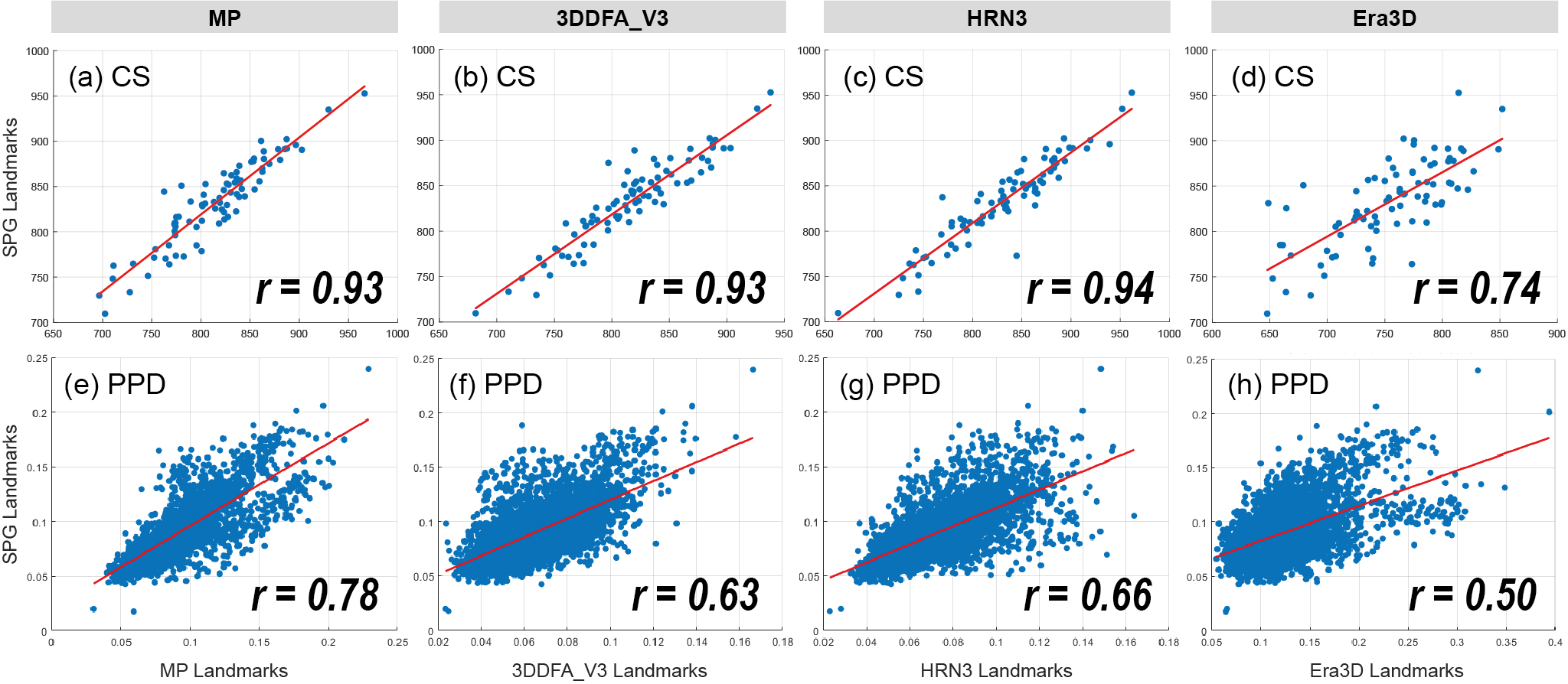}
    \end{center}
    \vspace{-0.5cm}
    \caption{(a-d) Centroid Size (CS) and (e-h) Pairwise Procrustes Distances (PPD) between SPG and low-cost methods. The $r$ coefficient shows the correlation between variables.}
    \label{fig:vm}
\end{figure}

Table \ref{tab:p2p} shows the point-to-point distances between the low-cost reconstruction methods with respect to SPG. MP is the method that offers the lowest average distance (<1mm). 3DDFA\_V3 and HRN3 show a similar behaviour, with the latter generating a slightly lower error. Era3D is the furthest away from the gold-standard.

\begin{table}[h]
\centering
\renewcommand{\arraystretch}{1.2} % Aumenta el espacio vertical
\setlength{\tabcolsep}{10pt} % Aumenta el espacio horizontal entre columnas (ajusta según necesidad)
\caption{Point-to-point (in mm) average distance $\pm$ standard deviation (SD) and maximum distance between low-cost methods and SPG.}
\begin{tabular}{c|c|c}
\hline
\textbf{Method} & \textbf{\begin{tabular}[c]{@{}c@{}}Avg. Distance $\pm$ SD (mm)\end{tabular}} & 
\textbf{\begin{tabular}[c]{@{}c@{}}Max. Distance (mm)\end{tabular}} \\ \hline
MP & 0.96 $\pm$ 0.75 & 6.32 \\ \hline
3DDFA\_V3 & 1.91 $\pm$ 1.44 & 11.57 \\ \hline
HRN3 & 1.45 $\pm$ 1.14 & 12.16 \\ \hline
Era3D & 2.82 $\pm$ 2.32 & 12.65 \\ \hline
\end{tabular}
\label{tab:p2p}
\end{table}

These findings are complemented by the surface-to-surface deviation coded in Figure \ref{fig:surface2surface}. We observed the loss of high-frequency detail in MP, especially in the commissures of the mouth and eyes. 3DDFA\_V3 showed a large deviation in the area of the nose, cheekbones and eyes. HRN3 preserved eye detail but deviated from SPG, though it improved in the mouth region. Era3D shows a flattening of the facial morphology and a large overall deviation from SPG.

%In general terms, all low-cost methods share a certain similarity with SPG reconstruction, with the exception of Era3D. MP reconstructions exhibit the highest visual resemblance to SPG, preserving key facial morphological features but lacking high-frequency details, especially in the eyes and mouth.

%As for 3DDFA\_V3 and HRN3, features such as the nose, mouth, and eyes show consistent similarity across all reconstructions but deviate somewhat from SPG, especially the first. 3DDFA\_V3 has difficulties in encoding the shape of the nose. Both seem to capture more high-frequency detail than MP. 

%Era3D reconstructions capture the most prominent morphological variations, such as the nose, but it is unable to capture high-frequency details and produces a flattened face shape.

\begin{figure} [t]
    \begin{center}
    \includegraphics[scale=0.6]{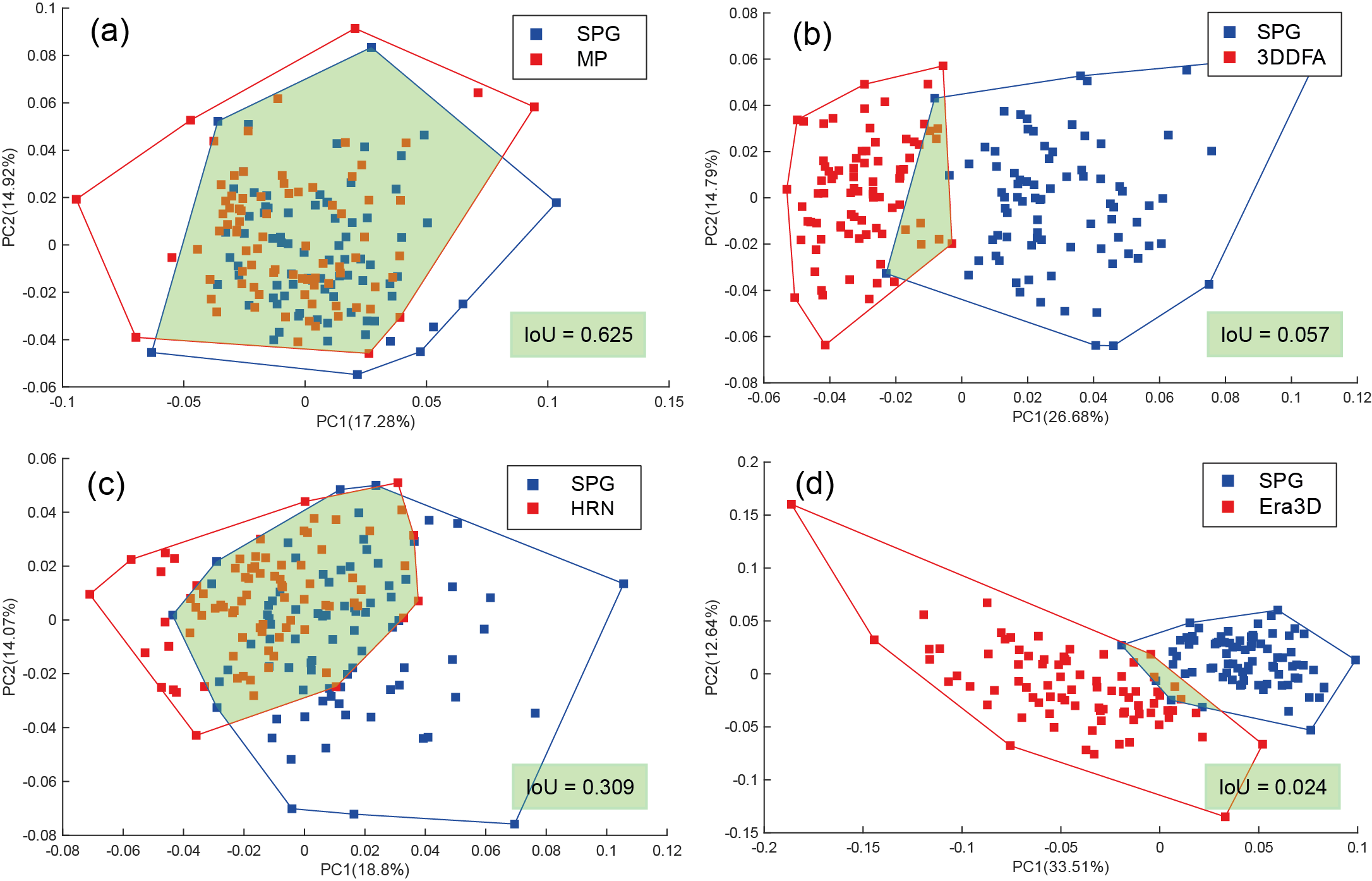}
    \end{center}
    \vspace{-0.5cm}
    \caption{GPA morphospaces comparing SPG with (a) MP, (b) 3DDFA\_V3, (c) HRN3 and (d) Era3D low-cost methods.}
    \label{fig:pca}
\end{figure}

\subsection{Morphometric evaluation}

%Geometric Morphometry (GM) landmarks-based evaluation is divided into two parts. First from the anatomical coordinates of the landmarks, and then by calculating feature vectors from these coordinates.

On the basis of the anatomical landmarks coordinates, the morphometric variables (Figure \ref{fig:vm}) display the correlation between the CS and PPD computed from the SPG and low-cost methods registered landmarks, portraying the correlation coefficient $r$ for each comparison. In all cases, the correlations were significant with a p-value $< 0.0001$. For both variables, we found a similar behaviour between the MP, 3DDFA\_V3 and HRN3 methods, while Era3D displayed lower correlation values. In the case of PPD we observed a higher correlation between the landmarks on the SPG and MP models (Figure \ref{fig:vm}e), which reveals a greater morphological similarity between the two methods.

From the landmark vectors resulting from the GPA, Figure \ref{fig:pca} shows the Intersection-over-Union (IoU) between the convex hulls of the high and low-cost groups in the PCA morphospace. The highest value (IoU = 0.62) was observed in the comparison of SPG and MP (Figure \ref{fig:pca}a), where the convex-hulls are very similar and defined by the same individuals, showing the morphological similarity between methods. The rest of the low-cost reconstructions displayed lower IoU values and very disparate convex hulls, demonstrating large deviations in facial shape from the gold-standard.

These results were accompanied by the Procrustes Distance (PD), where MP has the lowest value (PD = 0.026), followed by HRN3 (PD = 0.033), 3DDFA\_V3 (PD = 0.059) and Era3D (PD = 0.091) versus SPG GPA coordinates. However, for all methods, we found significant differences (p-value $<0.05$) between GPA coordinates of SPG and low-cost methods.

\begin{figure}[t]
    \begin{center}
    \includegraphics[scale=0.65]{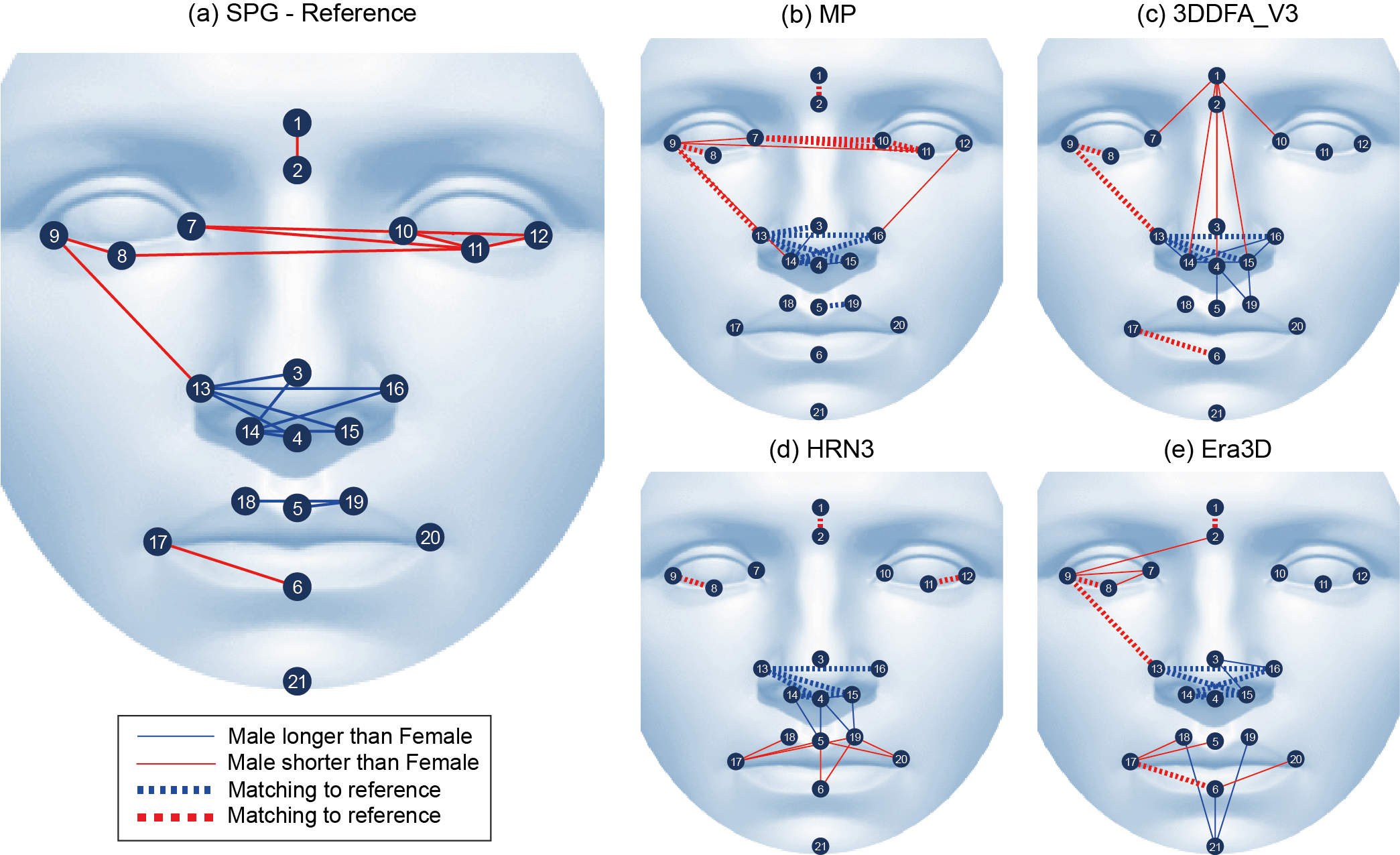}
    \end{center}
    \vspace{-0.5cm}
    \caption{EDMA top-10 significant inter-landmarks distances comparison between males and females for each 3D facial acquisition/reconstruction method: (a) SPG, used as reference, (b) MP, (c) 3DDFA\_V3 (d) HRN3 and (e) Era3D.}
    \label{fig:edma}
\end{figure}

Finally, Figure \ref{fig:edma} showcases the top-10 most significant shorter and longer inter-landmark distances between male and female groups for each low-cost method. Since the same individuals were used in all methods and no inter-observer landmark variability was introduced, the only variable factor was the acquistion/reconstruction method. This allowed us to directly compare which low-cost method best preserved the facial shape differences observed in the SPG gold-standard.

In the SPG reference case (Figure \ref{fig:edma}a), the most distinctive differences between males and females were in eye spacing and nose size. While all low-cost methods captured nose size differences, MP (Figure \ref{fig:edma}b) was the only method to also detect significant variations in eye spacing, making it the most accurate in replicating SPG’s sex-based shape differences. The matching inter-landmark distances across methods are represented as thick dashed lines in Figure \ref{fig:edma}.

These results are further supported by the matching distances (MD) values in Table \ref{tab:md}. In the top-5 comparison, MP achieved a perfect match (100\%) with the SPG reference, while the other methods showed lower coincidence. Even in the top-10 analysis, MP maintained the highest percentage of matching distances, reinforcing its higher similarity to SPG in consistency with the previous shape analysis experiments.

\begin{table}[t]
\centering
\renewcommand{\arraystretch}{1.2} % Aumenta el espacio vertical en las celdas
\caption{EDMA evaluation of low-cost methods: matching distances (\%) between the top-5 and top-10 significant inter-landmark distances.}
\begin{tabular}{c c c|c c|c c|c c}
\cline{2-9}
 & \multicolumn{8}{c}{\textbf{Top-5}} \\ \cline{2-9}
 & \multicolumn{2}{c|}{\textbf{MP}} & \multicolumn{2}{c|}{\textbf{3DDFA\_V3}} & \multicolumn{2}{c|}{\textbf{HRN3}} & \multicolumn{2}{c}{\textbf{Era3D}} \\ \cline{2-9}
 & \multicolumn{1}{c|}{Longer} & \multicolumn{1}{c|}{Shorter} & \multicolumn{1}{c|}{Longer} & \multicolumn{1}{c|}{Shorter} & \multicolumn{1}{c|}{Longer} & \multicolumn{1}{c|}{Shorter} & \multicolumn{1}{c|}{Longer} & Shorter \\ \hline
\multicolumn{1}{c|}{MD (\%)} & \multicolumn{1}{c|}{100} & \multicolumn{1}{c|}{100} & \multicolumn{1}{c|}{40} & \multicolumn{1}{c|}{40} & \multicolumn{1}{c|}{60} & \multicolumn{1}{c|}{40} & \multicolumn{1}{c|}{40} & 40 \\ \hline
\multicolumn{1}{c|}{Avg. MD (\%)} & \multicolumn{2}{c|}{100} & \multicolumn{2}{c|}{40} & \multicolumn{2}{c|}{50} & \multicolumn{2}{c}{40} \\ \hline
 & \multicolumn{8}{c}{\textbf{Top-10}} \\ \cline{2-9}
 & \multicolumn{2}{c|}{\textbf{MP}} & \multicolumn{2}{c|}{\textbf{3DDFA\_V3}} & \multicolumn{2}{c|}{\textbf{HRN3}} & \multicolumn{2}{c}{\textbf{Era3D}} \\ \cline{2-9}
 & \multicolumn{1}{c|}{Longer} & \multicolumn{1}{c|}{Shorter} & \multicolumn{1}{c|}{Longer} & \multicolumn{1}{c|}{Shorter} & \multicolumn{1}{c|}{Longer} & \multicolumn{1}{c|}{Shorter} & \multicolumn{1}{c|}{Longer} & Shorter \\ \hline
\multicolumn{1}{c|}{MD (\%)} & \multicolumn{1}{c|}{90} & \multicolumn{1}{c|}{50} & \multicolumn{1}{c|}{50} & \multicolumn{1}{c|}{40} & \multicolumn{1}{c|}{50} & \multicolumn{1}{c|}{30} & \multicolumn{1}{c|}{50} & 40 \\ \hline
\multicolumn{1}{c|}{Avg. MD (\%)} & \multicolumn{2}{c|}{70} & \multicolumn{2}{c|}{45} & \multicolumn{2}{c|}{40} & \multicolumn{2}{c}{45} \\ \hline
\end{tabular}
\label{tab:md}
\end{table}

%\subsection{Limitations}

\section{Conclusions}

This study has presented a novel methodology for evaluating low-cost 3D facial acquisition and reconstruction methods combining geometric and morphometrics accuracy metrics. By integrating GM-based shape analysis, the proposed methodology provides a statistically robust and biologically meaningful validation framework, offering a more comprehensive assessment of facial morphology preservation. %This approach enables a deeper evaluation of shape accuracy and anatomical consistency, which are essential for biomedical and clinical applications.

The methodology has been applied in a case of use comprising stereophotogrammetry as the gold standard, smartphone-based low-cost facial acquisition and deep learning-based 3D facial reconstruction methods from 2D images. 

Results demonstrate that smartphone scans show higher accuracy than facial reconstruction methods in terms of geometric and morphometric evaluation metrics. However, although MP showed better performance, its requirement for a static capture process presents practical limitations, which 2D imaging-based methods could overcome despite their slightly lower accuracy. These findings emphasize the need for further advancements in low-cost 3D reconstruction techniques to enhance their precision and reliability for clinical applications. The proposed methodology serves as a valuable tool for guiding these improvements.

\section{Acknowledgments}
We are grateful for the voluntary collaboration of all participants. This research was supported by AGAUR's Knowledge Industry Programme (2021 LLAV 00044), the Joan Oró grant from the DRU of the Generalitat de Catalunya (2024 FI-3 00160) and the European Social Fund, by the Fundación Álvaro Entrecanales-Lejeune predoctoral grant and by the Agencia Española de Investigación (PID2020-113609RB-C21/AEI/10.13039/501100011033, PID2023-147001OB-C22). We also acknowledge EuroHPC JU for awarding the project ID EHPC-043 access to MareNostrum5 at Barcelona Supercomputing Center and the Agència de Gestió d’Ajuts Universitaris i de Recerca (AGAUR) of the Generalitat de Catalunya (2021 SGR01396, 2021 SGR00706).

%
% ---- Bibliography ----
%
% BibTeX users should specify bibliography style 'splncs04'.
% References will then be sorted and formatted in the correct style.
%
% \bibliographystyle{splncs04}
% \bibliography{mybibliography}
%

\end{document}